# Shrinking the Inductive Programming Search Space with Instruction Subsets

Edward Mc Daid[1][a] and Sarah Mc Daid[1][b]
[1]Zoea Ltd., 20-22 Wenlock Road, London, N1 7GU, UK
{edward.mcdaid, sarah.mcdaid}@zoea.co.uk

Abstract: Inductive programming frequently relies on some form of search in order to identify candidate solutions. However, the size of the search space limits the use of inductive programming to the production of relatively small programs. If we could somehow correctly predict the subset of instructions required for a given problem then inductive programming would be more tractable. We will show that this can be achieved in a high percentage of cases. This paper presents a novel model of programming language instruction co-occurrence that was built to support search space partitioning in the Zoea distributed inductive programming system. This consists of a collection of intersecting instruction subsets derived from a large sample of open source code. Using the approach different parts of the search space can be explored in parallel. The number of subsets required does not grow linearly with the quantity of code used to produce them and a manageable number of subsets is sufficient to cover a high percentage of unseen code. This approach also significantly reduces the overall size of the search space - often by many orders of magnitude.

[a] https://orcid.org/0000-0001-8684-0868
[b] https://orcid.org/0000-0001-7643-6722

# 1 INTRODUCTION

The use of AI to assist in or even to automate the generation of computer software is an active area of research (e.g. Xu et. al. 2022, Nguyen and Nadi 2022). Many current systems are based on deep learning and recent work includes the use of large language models such as GPT-3 (Brown et. al. 2020). These involve training on large quantities of code although this can also raise concerns about transparency and ethics (Lemley and Casey 2021).

Work also continues on other approaches that are not traditionally associated with training (Cropper, Dumancic and Muggleton 2020, Petke et. al. 2018). Inductive programming aims to generate code directly from some form of specification – often in the form of input-output examples or test cases (Flener and Schmid 2008). Various techniques for inductive programming have been developed but fundamentally many of these rely on some form of search (Kitzelmann 2010).

Other than for trivial cases it is not possible to determine the outcome of a computation directly from source code without also executing the program. As a result some form of generate and test approach is unavoidable.

In computer programs a large number of language elements can be combined in many ways to quickly produce enormous numbers of candidate solutions. The size of the search space has limited inductive programming systems to the production of relatively small programs (Galwani et. al. 2015).

A major source of combinatorial growth in inductive programming is the number of instructions, comprising core language and standard library functions, and operators. This number varies by programming language but is frequently around 200 and can be more. It has been suggested that if we could somehow correctly predict the subset of instructions that were actually required for a given program then the problem of inductive programming would be easier and larger programs could be produced with given resources (McDaid 2019).

One way to produce slightly larger programs in a given time period is to distribute the work across many computers. In order to do this we need to partition the search space. Partitioning on the instruction set is attractive as most programs use a relatively small subset of instructions. In which case how do we define the subsets we will use?

This paper presents the results of a study that was carried out to inform the definition of instruction subsets for the Zoea inductive programming system (McDaid 2021). The initial sections in this paper describe our approach to the production and evaluation of instruction subsets. We then go on to discuss some significant findings that only became apparent once this work had been completed.

# 2 PRELIMINARIES

Let $C$ be a source code program in a high level imperative programming language $L$. $C$ is composed of one or more program units $U$ - corresponding to procedures, functions or methods. $L$ provides a set of instructions $IL$ comprising built in operators, core and standard library functions. Each $U$ contains a set of instructions $IU$ where $IU \subseteq IL \land IU \neq \emptyset$.

Given a collection of programs $SP1$ we can enumerate the corresponding family of program unit instruction subsets $SIU$ where $SIU \in P(IL) \land SIU \neq \emptyset$. (Here P refers to the power set.)

$IU$ is said to cover $U$ if $IU$ is the instruction subset for $U$ or $IU$ is the instruction subset for a different $U$ and a superset of the $IU$ for $U$. Each $IU$ trivially covers the corresponding $U$. We can also say that $SIU$ covers $SP1$. Coverage for a set of programs is quantified as the number of covered program units divided by the total number of program units expressed as a percentage.

Any $IU$ that is a subset of another $IU$ can be removed from $SIU$ without affecting the overall coverage of $SIU$ wrt $SP1$. Two or more $IU$s can be combined to form a new derived instruction subset $ID$. $ID$ provides the same aggregate coverage as its component $IU$s wrt $SP1$.

$SID$ is a family of instruction subsets comprising all $IU$s (that have not been removed or merged) union all $ID$s. $SID$ provides the same coverage as the original $SIU$ wrt $SP1$. We can enforce an upper limit $M1$ on $|IU|$ and an upper limit $M2$ on $|ID|$ during the creation of $SID$. Any $U$ where $|IU| > M1$ is silently ignored. $M2$ constrains which subsets of $SIU$ ($IU$s) can be combined to form $ID$s.

Any number of $ID$s can be created providing their respective component $IU$s are also removed and $|ID| <= M2$. Once created $SID$ can then be evaluated in terms of the coverage it provides wrt a different set of programs $SP2$.

# 3 APPROACH

## 3.1 Objectives

The primary goal of this work was to define a set of instruction subsets to support efficient clustering in the Zoea inductive programming system (McDaid 2021). Zoea employs a distributed blackboard architecture comprising many knowledge sources that operate in parallel. Knowledge source activations can also be partitioned according to instruction subsets.

Currently Zoea can efficiently utilise up to around 100 cores and it can create millions of activations in solving a problem. Efficiently utilising larger numbers of cores requires more fine-grained partitioning of the instruction set. This will enable Zoea to better leverage cloud-based deployment.

The number of subsets required to provide sufficient coverage for a wide variety of programs was unknown in advance. Based on experience in tuning the current system hundreds to thousands of subsets would likely result in acceptable performance.

This work evolved from and ultimately superseded an earlier effort to define random subsets, which is not described here in any detail. The strategy of deriving subsets from existing code was seen as a potential way to make subsets more representative of human originated software. It was also apparent that any such subsets would need to intersect with one another to some extent.

The target size of the instruction subsets is an important consideration. Smaller subsets make it possible to find programs with fewer instructions in a search space in less time. However, we also need to be able to produce programs with larger numbers of instructions. This suggests the use of multiple sets of instruction subsets of different sizes. In this scheme the subset size defines an upper limit on the number of unique instructions but not the size of the program that can be found. Operationally these can be used to either try to find programs with fewer unique instructions first or at the same time. The target subset sizes evaluated were 10 to 100 in increments of 10.

Input to the subset creation process is a set of non-empty program unit instruction subsets of various sizes. These are presented in the order in which they occur in the source code. No other associated metadata is provided. Often many of the input subsets are duplicates.

Depending on how much code is used there can be many program subsets and we would like to have a much smaller number of derived subsets. In broad terms this can be seen as a form of clustering. To enable program subsets to be clustered the derived subsets are allowed to be somewhat larger by a configurable amount. In producing derived instruction subsets of a given size we simply ignore any program subsets that are larger.

For a given derived subset size every program unit subset of that size or smaller will itself be a subset of at least one derived subset. This means that each of the largest program unit subsets will effectively form the core of one derived subset. The pigeonhole principle implies that there will be a minimum number of derived subsets required simply to accommodate the larger program subsets within them. This also suggests that we should process the larger program subsets first. If we process the input subsets in descending size order then at some stage the space remaining in the already partially populated derived subsets will be sufficient for smaller program subsets to be merged.

The performance of a set of derived instruction subsets can be measured by counting the number of unseen programs with an equal or smaller number of unique instructions that they cover. By unseen we simply mean programs that were not used for the production of the subsets. A program is covered by a set of subsets if all of its instructions are present in at least one of the subsets. This implies that any exhaustive search of the spaces defined by the subsets will eventually locate a covered program.

In summary, the key objectives were: to create sets of instruction subsets for different maximum subset sizes; to understand how many subsets are required for different subset maximum sizes; and determine the coverage for unseen code provided by the subsets created.

## 3.2 Method

Instruction subsets need to be representative of real world software, which implies they should also be derived from the same. A large quantity of code is required to ensure sufficient variety and ideally it should be the product of many different developers from a variety of contexts. The code also needs to be legally and ethically available.

GitHub (Microsoft 2022) was identified as a suitable source of software and it has been used in the past for similar analyses of code (Ray et. al. 2014). We used the largest 1,000 repositories on GitHub (as of 13 May 2022) and limited our analysis to Python (Martelli, Ravenscroft and Holden, 2017) programs only. Python was selected on the basis that it is a fairly popular language and the available instructions are representative of similar languages.

In each case the complete repository was downloaded as a zip file, extracted and non-Python files were discarded. Each Python program was split into program units (classes, methods, functions and mains) and tokenised using simple custom code based on regular expressions. Two of the repositories were excluded due to parsing errors. From the identified tokens the occurrences of each of a specific set of instructions were counted. Instructions that have no meaning in Zoea, such as those relating to variable assignment, were either mapped to an equivalent instruction if possible or else excluded. (E.g. '+=' is mapped to '+' while '=' is excluded.) No code or other information was used in any other way. The output of this step was a flat list of unique instruction names in alphabetical order for each program unit.

The analysis included 71,972 Python files containing 15,749,416 lines of code or 580,476,516 characters. From this 886,421 program units were identified. For each program unit the subset of instructions it contains was recorded. Many of the instruction subsets so identified were duplicates and when these were removed 345,120 unique instruction subsets remained.

During processing the instruction subsets were filtered to remove any that are proper subsets of another instruction subset. That is, if all of the instructions of subset A also occur in subset B then subset A is deleted. This process was repeated until no further subsets were deleted. This left 33,823 instruction subsets. These varied in size between 1 and 74 instructions (median: 3, standard deviation: 3.67). Many of the unique instruction subsets were very similar to one another, differing by only one or two instructions.

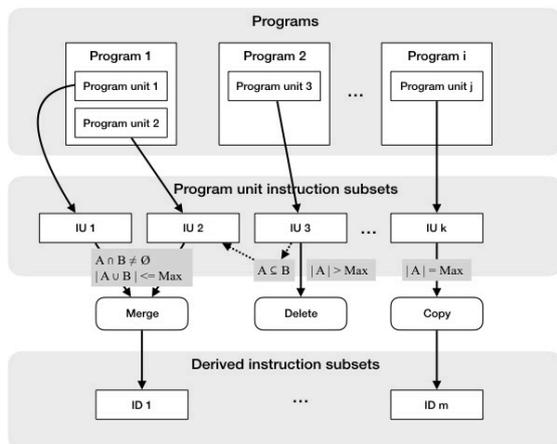

Figure 1: Overview of Subset Creation.

## 3.3 Clustering Algorithm

Figure 1 gives a conceptual overview of the subset creation process. Producing derived instruction subsets (IDs) from program unit instruction subsets (IUs) is essentially a clustering problem. A number of different clustering algorithms were developed and evaluated. The end result incorporates the two most successful of these together with some pre- and post-processing. Pseudo-code for the software is shown in Algorithm 1.

When a set of derived instruction subsets is created this is always with respect to a specified maximum subset size. This size limit also determines the target number of derived subsets as described in more detail later in this section.

```
delete any duplicate IUs
amplify IUs (see section 3.4)
delete IUs that are subsets of IU'
foreach IU do
  find Instr, Subsets where
    Instr is not a subset of IU and
    length(IU ∪ Instr) is maximum
  Let IU = IU ∪ { Instr }
  delete all IUs in Subsets
end foreach
create NumIDs empty IDs
foreach IU do
  if exists( empty ID ) then
    Let ID = IU
  else
    find IDs with max( | ID ∩ IU | )
    choose ID with min( | ID | )
    Let ID = ID ∪ IU
  end if
end foreach
merge IDs where size <= MaxIdSize
return set of IDs
```

Algorithm 1: Clustering algorithm.

Input subsets are processed in decreasing order of size. In order to improve performance all instruction subsets are kept internally sorted at all times.

Pre-processing involves de-duplication, amplification and subset removal. Amplification is described in the next section. Many input subsets are duplicates of which set one is retained. Any input subset that is wholly contained within another input subset is also removed.

The first clustering stage attempts to subsume the largest number of near subsets by adding a single additional instruction to each IU in turn. In choosing which instruction to add the algorithm determines how many other IUs will become subsets of the current IU if that instruction is added. The instruction that results in the removal of the greatest number of other IUs is selected.

The second clustering stage tries to merge each IU with the ID with which it has the greatest intersection. This involves pre-creating a specified number of empty IDs and then either populating the empty IDs or else merging the IU into the ID with both the largest intersection and the most remaining capacity. If all IDs are at their maximum capacity then additional empty IDs are created.

Post-processing involves merging any remaining small IUs and non-full IDs together. This is driven entirely by subset size and ignores similarity.

As noted already, the algorithm allows for the specification of both a maximum subset size and a target number of subsets. However, the number of subsets requested is not necessarily honoured if this proves to be impossible, in which case a minimal number of additional subsets are created. For each subset size the approximate number of subsets required is determined in advance by code that iterates over possible values in ascending order. The number of subsets required is detected when the number of subsets created equals the number of subsets requested. Given this requires considerable time the process has only been done in increments so the figures obtained are approximate, within the size of the increment.

### 3.4 Amplification

Initially, the process of merging subsets was simply seen as a way to reduce the number of derived subsets and to help standardize their sizes. However, it was observed that coverage against unseen code improved significantly after merging. This is partly because larger subsets provide better coverage than do small ones since they have a greater number of subsets. Also, combining different instruction subsets from unrelated code has similarities to software composition. Merging input subsets introduces additional intra-subset instruction co-occurrence variety that would not otherwise be present. It is interesting to note that adding an equivalent number of random instructions rather than merging does not give any detectable benefit.

In order to take greater advantage of this phenomenon an amplification step was introduced where smaller input subsets are merged with one another before clustering to create additional artificial program subsets. This was not explored systematically but the benefits do not seem to continue to accrue beyond a 50% increase in the number of subsets. Further work in this area may be fruitful.

## 4 RESULTS

### 4.1 Input Data

Figure 2 shows the size frequency distribution of the program unit instruction subsets. This contains both the numbers of subsets of different sizes and the cumulative percentage of the number of program units for each subset size. From this it can be seen that around 90% of program units have instruction subsets containing 10 or fewer unique instructions. Also, only around 2% of program unit subsets contain 20 or more instructions.

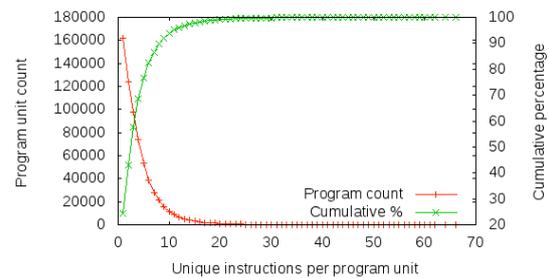

Figure 2: Program Unit Instruction Subset Size Frequency Distribution.

Figure 3 gives the frequency distribution for instructions across all of the code used. Here instructions are ranked in order of descending frequency. This shows that a small number of instructions are used very frequently and that many instructions are seldom used. This is similar to a Zipfian distribution that is often associated with human and artificial languages (Louridas, Spinellis and Vlachos 2008).

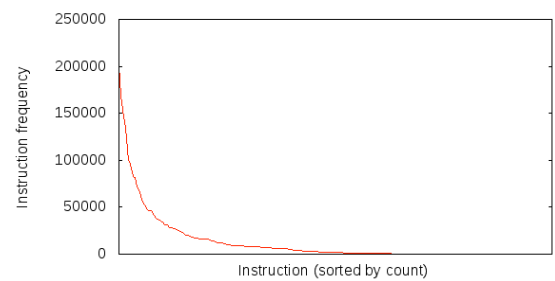

Figure 3: Instruction Frequency Distribution.

The ranked frequency distribution for co-occurring instruction pairs is similar to that for instructions although it is more pronounced. A much smaller percentage of instruction pairs co-occur frequently while the vast majority occurs infrequently or not at all. Again the tail is long and thin.

## 4.2 Coverage of Unseen Code

By definition the derived subsets will always give 100% coverage of the code that was used to create them. In other words, all of the instructions in each program unit instruction subset will be found together in at least one single derived instruction subset. However, this is not the purpose for which the derived subsets are created.

To be useful the derived subsets should also provide a high level of coverage for code that was not used for their creation. The level of coverage for unseen code was determined by nominating a percentage of the input code as a training set from which the derived subsets were then produced. The derived subsets that were produced in this way could then either be tested against a different section of the input codebase or else all of it.

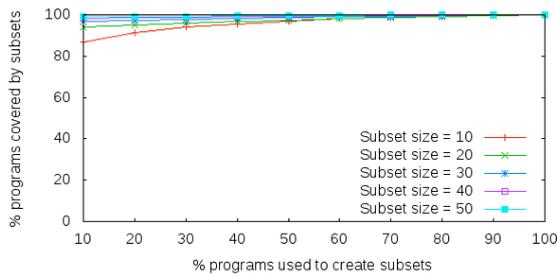

Figure 4: Unseen Code Coverage for Different Training Set Percentages and Subset Sizes.

To understand the relative coverage that was achieved using training sets of different sizes, subsets were produced using 10% to 100% of the available code in 10% increments. In each case the derived subsets were then evaluated against the entire codebase. These results are shown in Figure 4. These tests were executed over 50 times and it was clear early on that subsets produced using a relatively small percentage of the codebase can provide high coverage. For subset size 10 just 1% of the code produces subsets that provide 77.31% coverage.

Other tests - that are not reported here - were carried out to ensure the particular section of code from which the training set was taken had no adverse impact on the results.

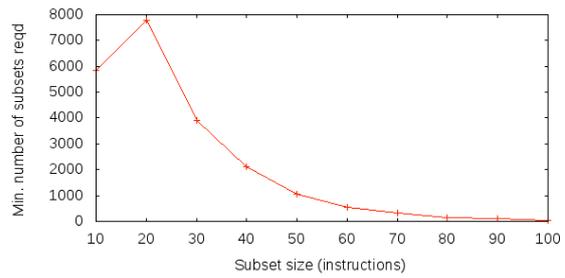

Figure 5: Number of Subsets Required by Subset Size.

## 4.3 Number of Subsets

As we have already noted the number of derived subsets required depends to a large extent on the maximum derived subset size. Figure 5 shows the numbers of derived subsets for various maximum sizes. Generally, the number of subsets reduces exponentially with increasing maximum subset size. That the number for maximum size 10 is less than 20 is probably due to the fixed size of the derived subset headroom in combination with the skewed subset size distribution.

The numbers of derived subsets are within what we would consider acceptable in order to support Zoea clustering. It is possible to reduce the number of subsets further by various means although these will also reduce some of the coverage for unseen code. For example, some of the subsets are redundant when considered solely in terms of coverage of the training set. In addition, many individual instructions can be removed from subsets on the same basis.

Another important factor is how the number of required subsets grows as the size of the training set is increased. This growth is not linear but instead decreases with each increment. The decreasing rate of growth suggests that subset size eventually stabilizes rather than growing indefinitely.

## 4.4 Search Space Reduction

The original motivation for using instruction subsets is to distribute work across many worker nodes in a cluster. An additional benefit which was not anticipated at the outset is that the size of the search space is also significantly reduced. It is easy to see why this is the case.

The search space for code approximates to a tree of a given depth with a branching factor largely determined by the number of instructions. Various approaches have been published for estimating the size of such a search tree (Kilby et. al. 2006). However, the reality in inductive programming is a little more complicated. Different instructions have different numbers of arguments and the data flows may span any number of levels forming a graph rather than a tree.

A more accurate estimate of cumulative search space size can be achieved by considering the number of values generated as successive layers of instructions are added. Inputs are considered to exist at level zero. If all instructions are applied at each level then single argument instructions must take their input from the previous level whereas two argument instructions only require one value from the previous level and another from any level. In this approach the number of search space nodes at a given level is the current total number of values excluding inputs.

Figure 6 shows the impact of different subset sizes on the size of the search space. This shows very large reductions in search space size – particularly for subsets of size 10. This is largely due to the reduction in branching factor from around 200 to 10. Note that the results in Figure 6 have already been multiplied by the corresponding number of subsets, although this makes little difference to the relative scale of the results.

It is clear that distributing the work across a number of nodes in this manner does not just enable the work to be completed more quickly. It also effectively reduces the overall amount of work that needs to be done.

If we didn't know any better then we might assume that human developers use all possible instruction subsets when they code. The results show that this is not the case. As we have seen the distribution of instructions that human coders use is heavily skewed and the distribution of pairs of instructions is even more so. This means that the occurrence of the majority of instruction pairs is vanishingly small in human originated code. The same must also apply to the distribution of instruction pairs in any instruction subsets derived from code.

We might also assume that all instruction subsets are required in order to write any possible piece of code. It is unknown whether or not this is actually the case. There are often many different ways of writing functionally equivalent programs using different subsets of instructions.

In any event people seem to get by using a relatively small proportion of all possible instruction subsets. This means there are a great many subsets that are not used very often - if at all. Every instruction subset, that does not include all instructions, accounts for a different and somewhat overlapping subset of the complete search space. Effectively it is the instruction subsets that are not used that account for the reduction in the size of the search space.

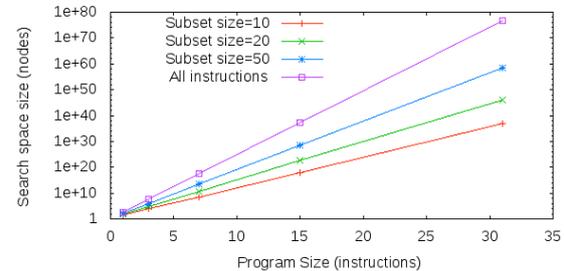

Figure 6: Search Space Reduction for Different Instruction Subset Sizes.

## 4.5 Subset Overlap

Instruction subsets often overlap. This is intentional and reflects the fact that instructions used in different program units frequently intersect. It is also partly due to the clustering process. The median overlap for subset size 10 is around 20% and for size 50 is around 40%. As a result there might be a concern that an excessive amount of effort may be wasted when using the subsets to partition work.

This is only true if the programs we are searching for are very small. However, as the size of the search space grows with increasingly longer programs the influence of instructions that are shared between subsets diminishes very quickly as shown in Figure 7.

Estimation of duplicate effort uses the same approach outlined earlier for search space size. Duplicated work at a given level corresponds to the size of the search space subtree for overlapping instructions only, divided by the size of the tree for the subset size number of instructions.

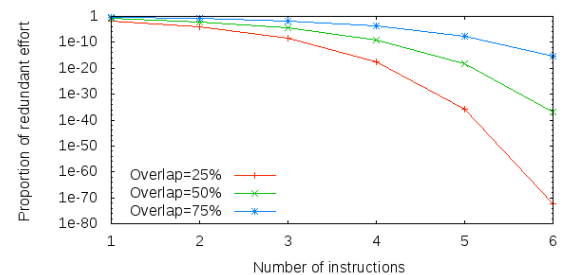

Figure 7: Proportion of Redundant Activity for Different Subset Overlap Percentages.

As the search tree grows, any values that have been produced using any non-duplicated instruction are distinct. Thus the proportion of values at each level that are produced exclusively from duplicated instructions quickly becomes insignificant. Similar duplication of effort is often tolerated in other techniques such as iterative deepening depth first search where the relative overhead is seen as marginal.

## 5 DISCUSSION AND FUTURE WORK

The frequency distribution of individual instructions can be seen as a fundamental form of software development knowledge. This distribution is highly skewed yet it is not altogether clear why this is the case or whether it has to be this way. Certainly many software developers would be aware of it, at least on reflection. Yet it does not seem to be a topic that has attracted much scientific attention.

Similarly, the co-occurrence of instructions in instruction subsets can be viewed as another very basic form of coding knowledge. Given that this has the potential to significantly reduce the size of the search space it can be viewed as an important form of tacit knowledge. Nevertheless this topic has received even less attention. This is not entirely surprising as superficially it seems like an unpromising line of enquiry – given the numbers of subsets involved and the degree of overlap between them.

In conducting this work it has become clear that there is a key trade-off between clustering and merging/amplification. While it is certainly possible to produce many fewer subsets through more aggressive clustering this comes at the price of less coverage for unseen code. We do not claim to have identified the optimum position on this continuum and more work in this area would be interesting. However, the current results are sufficient to support the initial operational deployment of this approach in Zoea - which is currently on-going.

By conceding that candidate solutions will only come from defined subsets of instructions we are accepting a compromise. We are willing to take any solution, potentially produced much faster, but there may be a small percentage of cases for which this approach might not succeed. More work will be required to quantify operational success in terms of generated solutions that meet the specification.

Other approaches to producing instruction subsets and alternative clustering algorithms are possible. Some of these may produce smaller sets of subsets and/or deliver greater coverage.

It is assumed that the results would also hold for other imperative programming languages. Most mainstream languages are very similar at the instruction level although the names and availability of some instructions can vary. It is unknown whether the approach would be beneficial in markedly different software development paradigms such as logic programming.

It is also assumed that the code sample used is representative of other code. The code used came from a single if large and popular source. In mitigation a large sample from many different repositories was used. However, code from smaller repositories and from other hosting sites may give different results.

Some instructions occur very frequently in the instruction subsets. One option would be to remove the most frequent instructions from the subsets and simply assume they always apply. This would have a dramatic effect on the size and number of the subsets. However, there is no clear boundary that separates frequent from infrequent instructions.

It is worth noting that instruction subsets are capable of generating many more programs than those from which they were derived. Also, lack of coverage does not mean that an equivalent program cannot be produced. There are many different ways to produce a functionally equivalent program – sometimes using different instructions.

Some of the individual subsets provide much more program coverage than others. This information could be used to prioritise the assignment of cluster jobs to increase the probability that a solution is found early.

This approach should be useful in any problem that involves searching a program configuration space. Integration should be a simple matter in any software that utilises a defined list of instructions. It is also worth considering whether a similar approach might be useful in domains beyond inductive programming. Combinatorial problems are common and the partitioning of work to better utilise large numbers of cores is a priority as well as being difficult.

The current work considers subset construction as an offline activity. In operational deployment this could alternatively be a continuous process. Similarly, the current focus is on subsets corresponding to human originated code. In the future machine generated code could be more abundant and it may be that instruction subsets derived from that would be different. In any event it makes no difference to Zoea where the subsets come from or how they are created.

It has long been suggested that neural and symbolic approaches to AI should be integrated (Kautz 2022). This will also require changes to how these technologies are used. The instruction subset approach isn't deep learning but is a step in that direction. While we have used the term 'training' in this paper, it is really an exercise in the elicitation of largely tacit knowledge. The data extracted from the code sample is both generic and very sparse. It also undergoes very little processing. The instruction subsets are completely human readable and as a result the whole process is transparent. What the approach does share with deep learning is the characteristic that rules are not written by people but rather produced mechanically and en masse.

Finally, it is a little surprising that there was anything new to learn in this area. In a way it shows the extent to which we still don't really understand what software is.

## 6 ETHICAL CONSIDERATIONS

The information extracted from the input source code was limited to the co-occurrence of instructions within each program unit. This is a flat list of the instruction names in alphabetic order.

It is important to note that in most cases these lists of instructions are not unique to the code they came from. Instead they are exceptionally common both as literal copies and also as subsets of one another. The use of instruction subsets cannot be considered a rights violation otherwise virtually all code in existence would already be in breach.

None of the information extracted represents or contains any executable code or information about any program, algorithm or fragment thereof. Also, there is nothing about the way in which this information will be used that could cause any similar or identical code to be generated, other than by pure chance. As such there can be no sense in which the copyright or intellectual property of the respective authors or the terms of any license have been violated.

## 7 CONCLUSIONS

We have outlined a technique for partitioning the search space of inductive programming using instruction subsets. This approach enables us to distribute inductive programming work across many computer cores by assigning each a distinct but overlapping subset of instructions. Testing suggests that the subsets generalise quickly, particularly when they are merged. This indicates that they should continue to work well even for completely different code. We have also shown that the use of instruction subsets reduces the size of the search space, in some cases by many orders of magnitude. Finally, we have demonstrated that the degree of duplication of effort as a result of overlapping instruction subsets quickly becomes insignificant as program size increases. We also believe our approach to be ethical in the sense that it does not exploit the intellectual property of open source developers.

## ACKNOWLEDGEMENTS

This work was supported entirely by Zoea Ltd. Zoea is a trademark of Zoea Ltd. Any other trademarks are the property of their respective owners. The authors acknowledge and respect the rights of open source developers.